\documentclass[10pt,twocolumn,letterpaper]{article}

\usepackage{cvpr}
\usepackage{times}
\usepackage{epsfig}
\usepackage{graphicx}
\usepackage{amsmath}
\usepackage{amssymb}
\usepackage{multirow}
\usepackage{booktabs}

\usepackage[breaklinks=true,bookmarks=false]{hyperref}

\cvprfinalcopy % *** Uncomment this line for the final submission

\def\cvprPaperID{****} % *** Enter the CVPR Paper ID here

\begin{document}

%%%%%%%%% TITLE
\title{Neighbourhood Watch: Referring Expression Comprehension via \\Language-guided Graph Attention Networks}

\author{Peng Wang$^{1}$\\
%{\tt\small peng.wang@adelaide.edu.au}
\and
Qi Wu$^{1}$\\
%{\tt\small qi.wu01@adelaide.edu.au}
\and
Jiewei Cao$^{1}$\\
%{\tt\small qi.wu01@adelaide.edu.au}
\and
Chunhua Shen$^{1}$ \\
%{\tt\small chunhua.shen@adelaide.edu.au,}
\and 
Lianli Gao$^{2}$\\
%{\tt\small juana.alian@gmail.com}
\and 
Anton van den Hengel$^{1}$\\
%{\tt\small anton.vandenhengel@adelaide.edu.au}\\
$^1$The University of Adelaide \hspace{.2cm} \\
$^2$The University of Electronic Science and Technology of China \\
{\tt\small \{peng.wang, qi.wu01, jiewei.cao, chunhua.shen, anton.vandenhengel\}@adelaide.edu.au}
}
\maketitle
%\thispagestyle{empty}

%%%%%%%%% ABSTRACT
\begin{abstract}
%This paper proposes to improve referring expression comprehension where the aim is to localize the object instance in an image described by the natural language description termed referring expression.
The task in referring expression comprehension is to localise the object instance in an image described by a referring expression phrased in natural language. As a language-to-vision matching task, the key to this problem is to learn a discriminative object feature that can adapt to the expression used. To avoid ambiguity, the expression normally tends to describe not only the properties of the referent itself, but also its relationships to its neighbourhood. 
%A So instead of learning object feature independently, in this paper, we propose to build a graph over the objects to model the object dependencies and design a language-guided graph attention mechanism to highlight the content relevant to the expression.
To capture and exploit this important information we propose a graph-based, language-guided attention mechanism. Being composed of \emph{node attention component} and \emph{edge attention component}, the proposed graph attention mechanism explicitly represents inter-object relationships, and properties with a flexibility and power impossible with competing approaches.
%To capture and exploit this important information we propose a graph-based, language-guided attention mechanism.  The graph explicitly represents inter-object relationships, and properties with a flexibility and power impossible with competing approaches.
%In this paper, we  propose to use the expression to inform object feature learning as a guidance to visual attention.
%Given an unambiguous expression tends to describe not only the referent itself but its relationships to one or multiple other objects as supporting clues,
%The graph attention is composed of two main components: a \emph{node attention} component to highlight relevant objects and an \emph{edge attention} component to identify the object relationships present in the expression. By summarizing the attended sub-graph centred by a potential object of interest, we can dynamically enrich the representation of this object so that it can adapt to the expression to achieve better matching performance. 
Furthermore, the proposed graph attention mechanism enables the comprehension decision to be visualisable and explainable. Experiments on three referring expression comprehension datasets show the advantage of the proposed approach.
\end{abstract}

%%%%%%%%% BODY TEXT
\section{Introduction}
\begin{figure}
\begin{center}
\includegraphics[scale=.37]{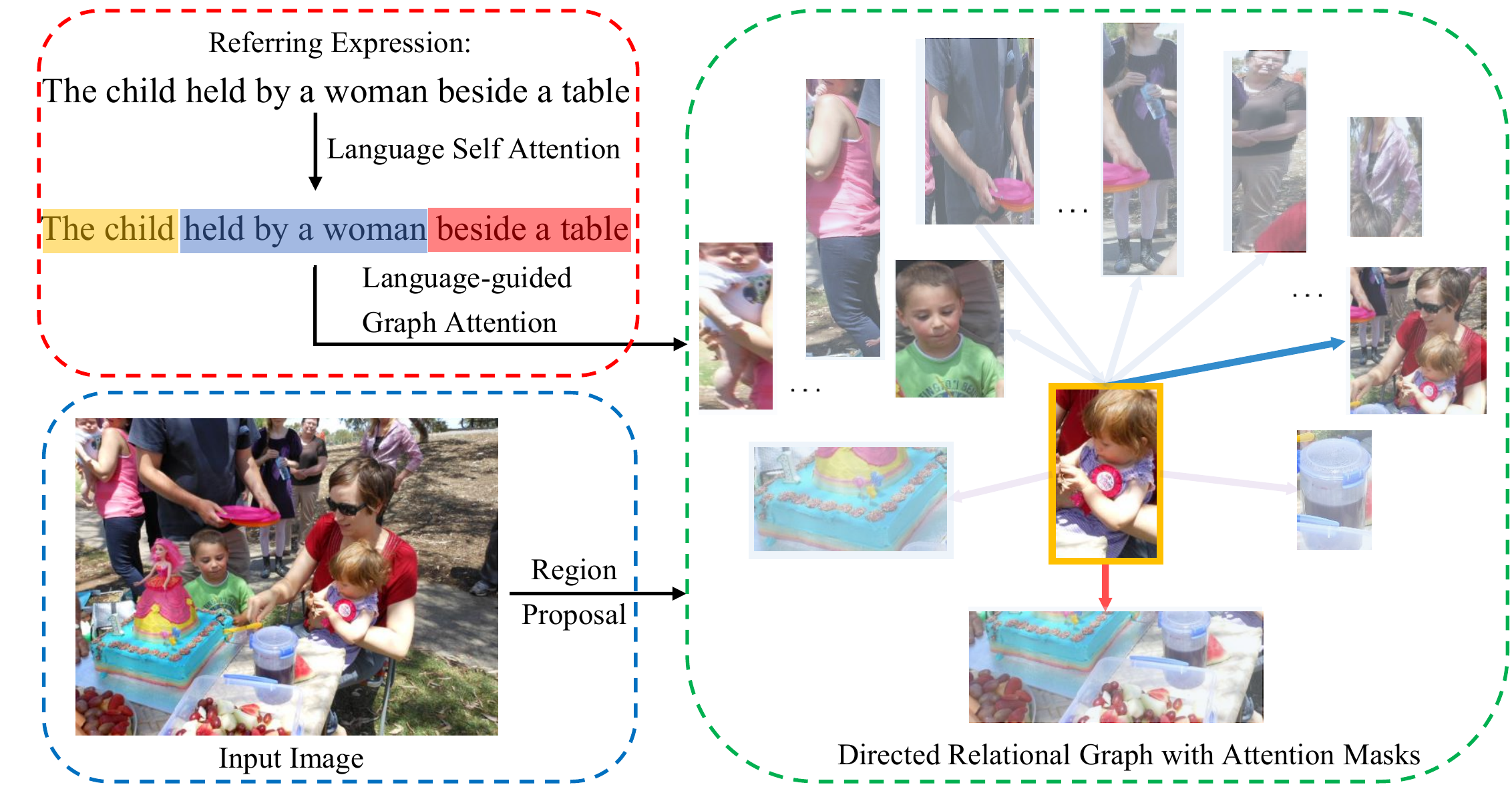}
\end{center}
\vspace{-0.3cm}
\caption{
%A Illustration of the graph attention mechanism for referring expression comprehension. 
A directed graph is built over the object instances of the image, where nodes correspond to object regions and edges (partially visualised) represent relationships between objects (blue and red edges denote intra- and inter-class relationships respectively). Graph attention predicts the attention distribution over the nodes as well as the edges, based on the decomposed information present in the expression. Summarising the attended object and its highlighted neighbours enables more discriminative feature.
Higher transparency here denotes a lower attention value. }
\vspace{-0.5cm}
\label{fig:intro}
\end{figure}
A referring expression is a natural language phrase that refers to a particular object visible in an image. 
%A In contrast to object detection in computer vision, where the task is to detect objects of interest, that are pre-defined by a set of isolated atom labels, r
Referring expression comprehension thus requires to identify the unique object of interest, referred to by the language expression~\cite{yu2017refexpr}. The critical challenge is thus the joint understanding of the textual and visual domains.

Referring expression comprehension can be formulated as a language-to-region matching problem, where the region with highest matching score is selected as the prediction.  Learning a discriminative region representation that can adapt to the language expression is thus critical. The predominant approaches \cite{Hu2016NaturalLO,mao2016generation,nagaraja16refexp} tend to represent the region by stacking various types of features, such as CNN features, spatial features or heuristic contextual features, and employ a LSTM to process the expression simply as a series of words. However, these approaches are limited by the monolithic vector representations that ignore the complex structures in the compound language expression as well as in the image.
Another potential problem for these approaches, and for more advanced modular schemes~\cite{Hu2017LearningTR,yu2018mattnet}, is that the language and region features are learned or designed independently without being informed by each other, which makes the features of the two modalities difficult to adapt to each other, especially when the expression is complex. Co-attention mechanisms are employed in~\cite{Deng_2018_CVPR, Zhuang_2018_CVPR} to extract more informative features from both the language and the image to achieve better matching performance. These approaches, however, treat the objects in the image in isolation and thus fail to model the relationships between them.  These relationships are naturally important in identifying the referent, especially when the expression is compound. For example, in Fig.~\ref{fig:intro}, the expression ``the child held by a woman beside a table'' describes not only the child but her relationships with another person and the table. In cases like this, focusing on the properties of the object only is not enough to localise the correct referent but we need to watch the neighbourhood to identify more useful clues.

To address the aforementioned problems, we propose to build a \textbf{directed graph} over the object regions of an image to \textbf{model the relationships between objects}. In this graph the nodes correspond to the objects and the edges represent the relationships between objects. 
%The candidate object regions are provided as ground truth or obtained by a region proposer (such as Faster-RCNN), depending on the experiments setting. 
On top of the graph, we propose a \textbf{l}anguage-guided \textbf{gr}aph \textbf{a}ttention \textbf{n}etwork (\textbf{LGRAN}) to highlight the relevant content referred to by the expression. The graph attention is composed of two main components: a \emph{node attention} component to highlight relevant objects and an \emph{edge attention} component to identify the object relationships present in the expression. Furthermore, the edge attention is divided into intra-class edge attention and inter-class edge attention to distinguish relationships between objects of the same category and those crossing categories. Normally, these two types of relationships are different visually and semantically. The three types of attention are guided by three corresponding language parts which are identified within the expression through a self-attention mechanism \cite{Hu2017LearningTR,yu2018mattnet}. By summarising the attended sub-graph centred on a potential object of interest, we can dynamically enrich the representation of this object in order that it can better adapt to the expression, as illustrated in Fig.~\ref{fig:intro}.
%While the former tend to represent, \textit{size} (e.g., taller, bigger, smallest), \textit{location} (e.g., in the middle, between two other kids) or other intra-class relationships (e.g., hugging), the latter include more general relationships occurring between objects from different classes (e.g., riding horse, reading book). 

%Inspired by the self-attention based language decomposition in \cite{7780381,Hu2017LearningTR,yu2018mattnet}, we additionally adopt a self-attention mechanism to decompose the expression into three components: subjective, intra-class relationship and inter-class relationship. Each of these parts corresponds to one graph attention type and is used to guide them. Being attended by the expression, the graph can be viewed as a set of sub-graphs where each sub-graph is centered by an object. By summarizing the highlighted node feature as well as the intra- and inter-class edge features of the object, we can dynamically enrich the representation of the object of interest and consequently obtain more discriminative features that can better adapt to the expression, as illustrated in Fig.~\ref{fig:intro}.

Another benefit of the proposed graph attention mechanism is that it renders the referring expression decision both \textbf{visualisable} and \textbf{explainable}, because it is capable of grounding the referent and other supporting clues (\ie its relationships with other objects) onto the graph.
We conduct experiments on three referring expression datasets (RefCOCO, RefCOCO+ and RefCOCOg). The experimental results show the advantage of the proposed language-guided graph attention network. We outperform the previous best results on almost all splits, under different settings.

\section{Related Work}
\begin{figure*}[t]
\begin{center}
\includegraphics[scale=.57]{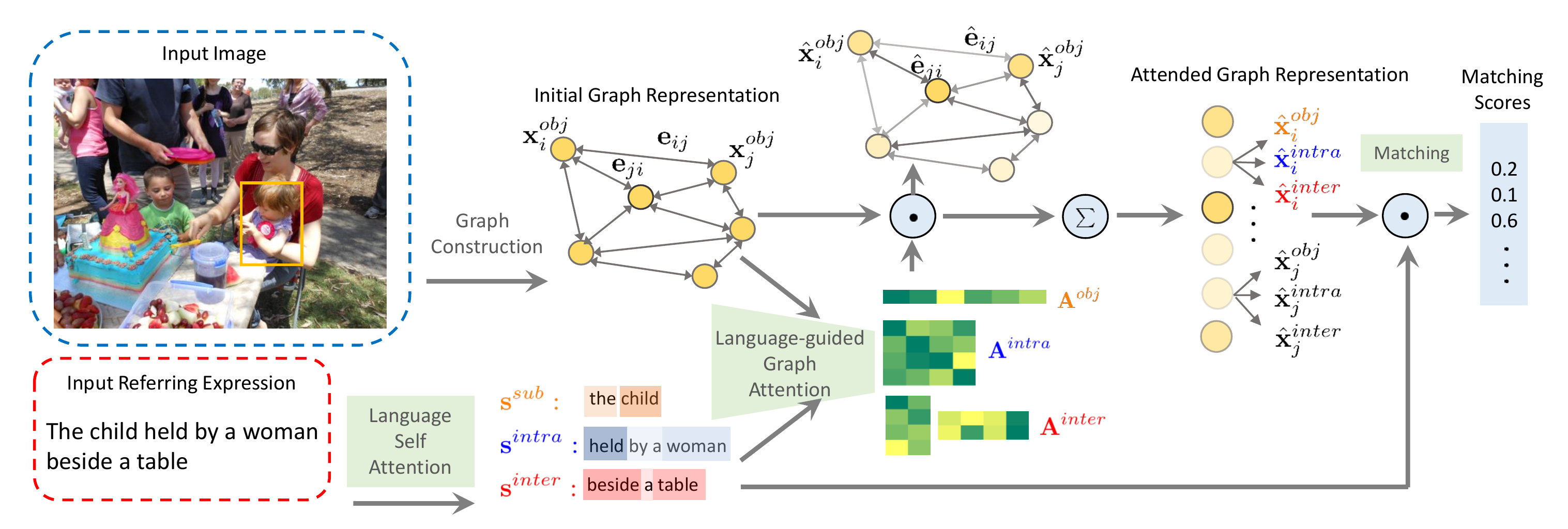}
\end{center}
\vspace{-0.3cm}
\caption{Overview of the proposed language-guided graph attention networks for referring expression comprehension. The network is composed of three modules: language-self attention module, language-guided graph attention module, and matching module.}
\vspace{-0.1cm}
\label{fig:flow}
\end{figure*}

\paragraph{Referring Expression Comprehension} Conventional referring expression comprehension is approached using a CNN/LSTM framework \cite{Hu2016NaturalLO,mao2016generation,nagaraja16refexp,unc_refexp}. The LSTM takes as input a region-level CNN feature and a word vector at each time step, and aims to maximize the likelihood of the expression given the referred region. These models incorporate contextual information visually, and how they achieve this is one of the major differentiators of the various approaches.  For example, the work in \cite{Hu2016NaturalLO} uses a whole-image CNN feature as the region context, the work in \cite{nagaraja16refexp} learns context regions via multiple-instance learning, and in~\cite{unc_refexp}, the authors use visual differences between objects to represent the visual context. 

Another line of work treats referring expression comprehension as a metric learning problem \cite{Luo2017,mao2016generation,reconstruction,Wang_2016_CVPR}, whereby the expression feature and the region feature are embedded into a common feature space to measure the compatibility. The focus of these approaches lies in how to define the matching loss function, such as softmax loss~\cite{Luo2017,reconstruction}, max-margin loss~\cite{Wang_2016_CVPR}, or Maximum Mutual Information (MMI) loss \cite{mao2016generation}. These approaches tend to use a single feature vector to represent the expression and the image region.  These monolithic features ignore the complex structures in the language as well in the image, however. To overcome this limitation of monolithic features, self-attention mechanisms have been used to decompose the expression into sub-components and learn separate features for each of the resulting parts \cite{Hu_2017_CVPR,yu2018mattnet,Zhang_2018_CVPR}. Another potential problem for the aforementioned methods is that the language and region features are learned independently without being informed by each other. To learn expression features and region features that can better adapt to each other, co-attention mechanisms have been used~\cite{Deng_2018_CVPR, Zhuang_2018_CVPR}. These methods process the objects in isolation, however, and thus fail to model the object dependencies, which are critical in identifying the referent. In our model, we build a directed graph over the object regions of an image to model the relationships between objects. On top of that, a language-guided graph attention mechanism is proposed to highlight the relevant content referred to by the expression.

\vspace{-15pt}
\paragraph{Graph Attention} In~\cite{velickovic2018graph} graph attention is applied to other graph-structured data, including document citation networks, and protein-protein interactions. The differences between their graph attention scheme and ours are threefold. First, their graph edges reflect the connections between nodes only, while ours additionally encode the relationships between objects (that have properties of their own). Second, their attention is obtained via self-attention or the interaction between nodes, but our attention is guided by the referring statement. Third, they update the node information as a weighted sum of the neighbouring representations, but we maintain different types of features to represent the node properties and node relationships. In terms of building a graph to capture the structure in the structural data, our work is also related to graph neural networks \cite{CNG,Jain_2016_CVPR,SituationsICCV17,grn,Teney_2017_CVPR}. Our focus in this paper is on identifying the expression-relevant information for an object for better language-to-region matching.
%Our approach is also related to graph network for other vision-to-language tasks, such as image captioning \cite{Yao_2018_ECCV} or VQA \cite{Teney_2017_CVPR}. Apart from different application scenarios, their works focus on learning a  image-level feature via message passing, but we intend to enrich the feature of a given object by inspecting the object properties and its relationships with others 

\section{Language-guided Graph Attention Networks (LGRANs)}
%The input are a referring expression $r$ and an image $I$ with an associated set of candidate objects $O=\{o_i\}^N_{i=1}$.

Here we elaborate on the proposed language-guided graph attention networks (LGRANs) for referring expression comprehension. Given the expression $r$ and an image $I$, the aim of referring expression comprehension is to localise the object $o^\star$ referred to by $r$ from the object set $\mathcal{O}=\{o_i\}^{N}_{i=1}$ of $I$. The candidate object set is given as ground truth or obtained by an object proposal generation method, such as region proposal network \cite{faster-rcnn}, depending on the experimental setting. We evaluate both cases in Sec.~\ref{sec:exp}.

%First we give the overview structure of the network, which is followed by the detailed illustration of the key modules.

As illustrated in Fig.~\ref{fig:flow}, LGRANs is composed of three modules: (1) the language self-attention module, which adopts a self-attention scheme to decompose the expression $r$ into three parts that describe the \textit{subject}, \textit{intra-class relationships} and \textit{inter-class relationships}, and learn the corresponding representations $\mathbf{s}^{sub}$, $\mathbf{s}^{intra}$ and $\mathbf{s}^{inter}$; (2) the language-guided graph attention module, which builds a directed graph over the candidate objects $\mathcal{O}$,  highlights the nodes (objects), intra-class edges (relationships between objects of the same category) and inter-class edges (relationships between objects from different categories) that are relevant to $r$ under the guidance of $\mathbf{s}^{sub}$, $\mathbf{s}^{intra}$ and $\mathbf{s}^{inter}$, and finally obtains three types of expression-relevant representations for each object; (3) the matching module, which computes the expression-to-object matching score. We now describe these modules in detail.

\subsection{Language Self-Attention Module}
\label{sec: lang_attn}

Languages are compound and monolithic vector representations (such as the output of a LSTM at the final state) ignore the rich structure in the language. Inspired by the idea of decomposing compound language into sub-structures in various vision-to-language tasks \cite{7780381,Hu2017LearningTR,Hu_2017_CVPR,yu2018mattnet}, we decompose the expression into sub-components as well. To fulfill their purpose referring expressions tend to describe not only the properties of the referent, but also its relationships with nearby objects. We thus decompose the expression $r$ into three parts: subject $r^{sub}$, intra-class relationship $r^{intra}$, and inter-class relationship $r^{inter}$.

\begin{figure}[t]
\begin{center}
\includegraphics[scale=.4]{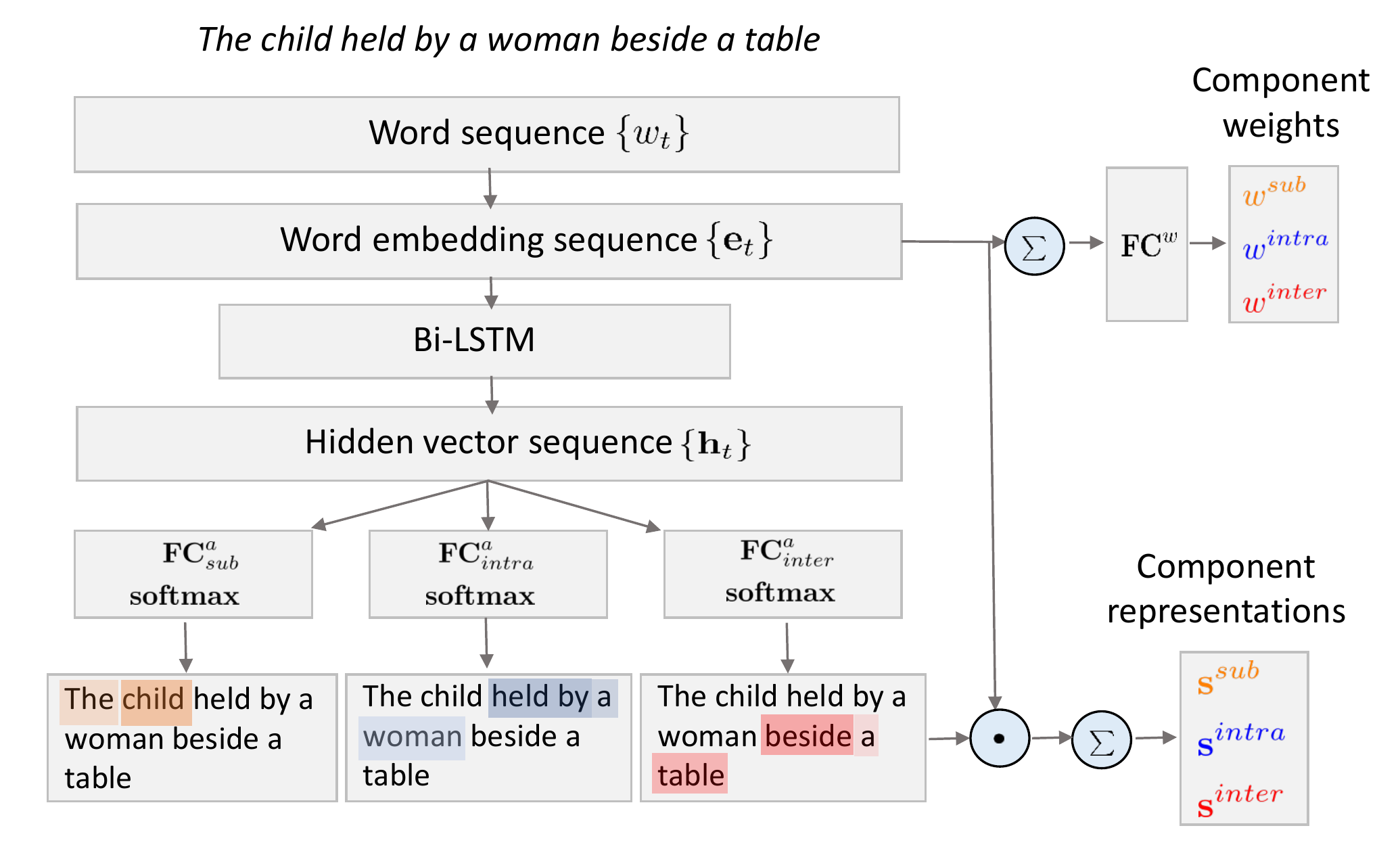}
\end{center}
\vspace{-0.6cm}
\caption{Illustration of the language self-attention module.}
\vspace{-0.4cm}
\label{fig:lang_attn}
\end{figure}

%where the former tend to represent, \textit{size} (e.g., taller, bigger, smallest), \textit{location} (e.g., in the middle, between two other kids) or other intra-class relationships (e.g., hugging), the latter include more general relationships occurring between objects from different classes (e.g., riding horse, reading book),

%The idea of decomposing the compositional language into sub-structures has been applied to various vision to language tasks \cite{7780381,Hu2017LearningTR,yu2018mattnet,Hu_2017_CVPR}. The decomposition is achieved either by off-the-shelf language parser \cite{7780381} or through the self-attention mechanisms \cite{Hu2017LearningTR,yu2018mattnet,Hu_2017_CVPR}. In order to learn the decomposition from the data, similar to \cite{Hu2017LearningTR,yu2018mattnet,Hu_2017_CVPR} we exploit the self-attention scheme. Fig.~\ref{fig:lang_attn} shows the basic idea.

There are mainly two language parsing approaches: off-the-shelf language parsers \cite{7780381} or self-attention \cite{Hu2017LearningTR,Hu_2017_CVPR,yu2018mattnet}. In this paper, we apply the self-attention scheme due to its better performance. Fig.~\ref{fig:lang_attn} shows the high-level idea of our language attention mechanism. Given an expression $r$ with $T$ words $r=\{w_t\}^T_{t=1}$, we first embed the words' one-hot representations into a continuous space $\{\mathbf{e}_t\}^T_{t=1}$ using a non-linear mapping function $f_e$. Then $\{\mathbf{e}_t\}$ are fed into a Bi-LSTM \cite{bi-lstm} to obtain a set of hidden state representations $\{\mathbf{h}_t\}^T_{t=1}$.
%\begin{align}
%\begin{split}
%&\overrightarrow{\mathbf{h}}_t=L\overrightarrow{ST}M(\overrightarrow{\mathbf{h}}_{t-1}),\\
%&\overleftarrow{\mathbf{h}}_t=L\overleftarrow{ST}M(\overleftarrow{\mathbf{h}}_{t+1}),\\
%&\mathbf{h}_t = [\overrightarrow{\mathbf{h}}_t, \overleftarrow{\mathbf{h}}_t].
%\end{split}
%\end{align}
Next, three individual fully-connected layers followed by softmax layers are applied to $\{\mathbf{h}_t\}$ to obtain three types of attention values, being subject attention $\{a^{sub}_t\}^T_{t=1}$, intra-class relationship attention $\{a^{intra}_t\}^T_{t=1}$ and inter-class relationship attention $\{a^{inter}_t\}^T_{t=1}$. As the attention values are obtained by the same way for all three components, for simplicity we only show the details for the calculation of the subject component $r^{sub}$.  Let
{\small\begin{align}
\vspace{-0.32cm}
a^{sub}_t=\frac{\exp(\mathbf{w}_{sub_a}^{\intercal}\mathbf{h}_t)}{\sum^T_{i=1}{\exp(\mathbf{w}_{sub_a}^{\intercal}\mathbf{h}_i)}},
\vspace{-0.3cm}
\end{align}}
where $\mathbf{w}_{sub_a}$ denotes $\textbf{FC}^a_{sub}$ in Fig.~\ref{fig:lang_attn}.
Then, the attention values are applied to the embedding vectors $\{\mathbf{e}_t\}$ to derive three representations: $\mathbf{s}^{sub}$, $\mathbf{s}^{intra}$ and $\mathbf{s}^{inter}$. Here we choose $\mathbf{s}^{sub}$ for illustration:
\begin{align}
\mathbf{s}^{sub} = \sum_{t=1}^T{a^{sub}_t\cdot{\mathbf{e}_t}}.
\vspace{-3pt}
\end{align}

Inspired by \cite{yu2018mattnet}, we apply another linear mapping $\text{FC}^w$ to the pooled embedding vector, $\mathbf{e}=\sum_{t=1}^{T}{\mathbf{e}_t}$, to derive three weights 
$[{w^{sub},w^{intra},w^{inter}}]$.  
%$[{\mathbf{w}^{sub},\mathbf{w}^{intra},\mathbf{w}^{inter}}]$.
These serve as the weights for $[r^{sub}, r^{intra}, r^{inter}]$ in expression-to-region matching, that will be introduced in Sec.~\ref{sec: matching}. Again we present how to obtain $w^{sub}$ only,
{\small
\begin{align}
%\nonumber
w^{sub}=\frac{\exp(\mathbf{w}_{sub_w}^{\intercal}\mathbf{e})}{\exp(\mathbf{w}_{sub_w}^{\top}\mathbf{e})+\exp(\mathbf{w}_{intra_w}^{\top}\mathbf{e})+\exp(\mathbf{w}_{inter_w}
^{\top}\mathbf{e})},
\end{align}}
where $\mathbf{w}_{sub_w}$, $\mathbf{w}_{intra_w}$, $\mathbf{w}_{inter_w}$ denote linear mappings.

\subsection{Language-guided Graph Attention Module}
The language-guided graph attention module is the key of the network. It builds a graph over the objects of an image to model object dependencies and identifies the nodes and edges relevant to the expression to dynamically learn object representations that adapt to the language expression.

\vspace{-5pt}
\subsubsection{Graph construction}
\label{sec: graph_construction}
Given the object or region set $\mathcal{O}=\{o_i\}^N_{i=1}$ of an image $I$, we build a directed graph $\mathcal{G}=\{\mathcal{V},\mathcal{E}\}$ over $\mathcal{O}$, where $\mathcal{V}=\{v_i\}^N_{i=1}$ is the node set and $\mathcal{E}=\{e_{ij}\}$ is the edge set. Each node $v_i$ corresponds to an object $o_i\in\{\mathcal{O}\}$ and an edge $e_{ij}$ denotes the relationship between $o_i$ and $o_j$. Based on whether the two nodes connected by an edge belong to the same category or not, we divide the edges into two sets: intra-class edges $\mathcal{E}^{intra}$ and inter-class edges $\mathcal{E}^{inter}$. That is, $\mathcal{E}=\mathcal{E}^{intra}\cup\mathcal{E}^{inter}$ and $\mathcal{E}^{intra}\cap\mathcal{E}^{inter}=\emptyset$. Assume $c(v_i)$ denotes the category of $v_i$, the two types of edges can be represented as,
%\begin{align}
$\mathcal{E}^{intra}=\{e_{ij}: c(v_i)=c(v_j)\}$ and
$\mathcal{E}^{inter}=\{e_{ij}: c(v_i)\neq{c(v_j)\}}$. 
%\end{align}
%We call an edge $e_{ij}$ intra-class edge if $e_{ij}\in{\mathcal{E}^{intra}}$ and otherwise inter-class edge if $e_{ij}\in{\mathcal{E}^{inter}}$.

Considering that an object typically only interacts with objects nearby, we define edges between an object and its neighbourhood. Specifically, given a node $v_i$, we rank the remaining objects of the same category, $\{v_j: c(v_j)=c(v_i)\}$, based on their distances to $v_i$ and define the intra-class neighbourhood $\mathcal{N}^{intra}_i$ of $v_i$ as the top $k$ ranked intra-class objects. Similarly, we define the inter-class neighbourhood $\mathcal{N}^{inter}_i$ of $v_i$ to be the top $k$ ranked objects that belong to other categories. For a node $v_i$, we define an edge between $v_i$ and $v_j$ if and only if $v_j \in \mathcal{N}^{intra}_i$ or $v_j\in\mathcal{N}^{inter}_i$. A bigger $k$ leads to a denser graph, and to balance the efficiency and representation capacity, we set $k=5$.

We extract two types of node features for each node $v_i$: appearance feature $\mathbf{v}_i$ and spatial feature $\mathbf{l}_i$. To obtain the appearance feature, we first resize the corresponding region $o_i$ to $224\times{224}$ and feed it to VGG16 net \cite{Simonyan14c}. The \textit{Conv5\_3} features $\mathbf{V}\in\mathrm{R}^{7\times{7}\times{512}}$ are pooled over the \textit{height} and \textit{width} dimensions to obtain the representation $\mathbf{v}_i\in\mathrm{R}^{512}$. The spatial feature $\mathbf{l}_i$ is obtained as in \cite{yu2017refexpr}, which is a 5-dimensional vector, encoding the top-left, bottom-right coordinates and the size of the bounding box with respect to the whole image, \ie, $\mathbf{l}_i=[\frac{x_{tl}}{W}, \frac{y_{tl}}{H}, \frac{x_{br}}{W}, \frac{y_{br}}{H}, \frac{w\cdot{h}}{W\cdot{H}}]$. The node representation is a concatenation of the appearance feature and spatial feature, \ie $\mathbf{x}^{obj}_i=[\mathbf{v}_i, \mathbf{l}_i]$. It has been shown that the relative spatial feature between two objects is a strong representation to encode their relationship \cite{Zhuang_2017_ICCV}. Similarly, we model the edge between $v_i$ and $v_j$ based on their relative spatial information. Suppose the centre coordinate, width and height of $v_i$ are represented as $[x_{c_i}, y_{c_i}, w_i, h_i]$, and the top-left coordinate, bottom-right coordinate, width and height of $v_j$ are represented as $[x_{tl_j},y_{tl_j},x_{br_j},y_{br_j},w_j,h_j]$, then the edge representation is represented as, $\mathbf{e}_{ij}=[\frac{x_{tl_j}-x_{c_i}}{w_i},\frac{y_{tl_j}-y_{c_i}}{h_i}, \frac{x_{br_j}-x_{c_i}}{w_i}, \frac{y_{br_j}-y_{c_i}}{h_i},\frac{w_j\cdot{h_j}}{w_i\cdot{h_i}}]$.

\vspace{-5pt}
\subsubsection{Language-guided Graph attention}
\label{sec: graph_attn}
The aim of graph attention is to highlight the nodes and edges that are relevant to the expression $r$ and consequently obtain object features that adapt to $r$. The graph attention is composed of two parts: \emph{node attention} and \emph{edge attention}. Furthermore, the edge attention can be divided into \emph{intra-class edge attention} and \emph{inter-class edge attention}.
Mathematically, this process can be expressed as,
{\small
\begin{align}
\{\mathbf{A}^{obj}, \mathbf{A}^{intra},\mathbf{A}^{inter}\} = f(\{\mathbf{x}^{obj}_i\}, \{\mathbf{e}_{i,j}\}, \mathbf{s}^{sub}, \mathbf{s}^{intra}, \mathbf{s}^{inter}),
%A *** This 'function' f() seems to return 3 things, which is something mathematical functions don't do.  Maybe put them in a set \{\}
\end{align}}
where $\mathbf{A}^{obj}$, $\mathbf{A}^{intra}$, and $\mathbf{A}^{inter}$ denote node attention values, intra-class edge attention values, and inter-class edge attention values respectively. The series of $\mathbf{s}$ are the attended features from the language part. The function $f$ is a graph attention mechanism that is guided by the language, which will be introduced as following three parts.

\vspace{-10pt}
\paragraph{The node attention} The node attention mechanism is inspired by the bottom-up attention \cite{Anderson_2018_CVPR}, which enables attention to be calculated at the level of objects and other salient image regions~\cite{Uijlings13,10.1007/978-3-319-10602-1_26}. Given the node features $\{\mathbf{x}^{obj}_i\}^N_{i=1}$, where $\mathbf{x}^{obj}_i=[\mathbf{v}_i,\mathbf{l}_i]$, and the subject feature $\mathbf{s}^{sub}$ of $r$ in Sec.~\ref{sec: lang_attn}, the node attention is computed as,
\begin{align}
\begin{split}
&\mathbf{v}^e_i = f^v_{emb}(\mathbf{v}_i)\\
&\mathbf{l}^e_i = f^l_{emb}(\mathbf{l}_i)\\
&\mathbf{x}^{e,obj}_i = [\mathbf{v}^e_i,\mathbf{l}^e_i]\\
&\mathbf{x}^{a,obj}_i=\tanh(\mathbf{W}^a_{s,sub}\mathbf{s}^{sub}+\mathbf{W}^a_{g,obj}\mathbf{x}^{e,obj}_i)\\
&{A^{obj}_i}' = \mathbf{w}^{\intercal}_{a,obj}\mathbf{x}^{a,obj}_i\\
&{A^{obj}_i}=\frac{\exp({A^{obj}_i}')}{\sum_j^N{\exp({A^{obj}_j}')}},
\end{split}
\label{equ:obj}
\end{align}
where $f^v_{emb}$ and $f^l_{emb}$ are MLPs used to encode appearance and local features of $v_i$ separately, $\mathbf{W}^a_{g,obj}$ and $\mathbf{W}^a_{s,sub}$ map the encoded node feature $\mathbf{x}^{e,obj}_i$ and subject feature $\mathbf{s}^{sub}$ of $r$ into vectors of the same dimensionality, $\mathbf{w}_{a,obj}$ calculates the attention values $\{{A^{obj}_i}'\}$ for $\{v_i\}$, and all these attention values $\{{A^{obj}_i}'\}^N_{i=1}$
are fed into a softmax layer to obtain the final attention values, $\mathbf{A}^{obj}=\{A^{obj}_i\}$.

\vspace{-12pt}
\paragraph{The intra-class edge attention} We obtain the attention values for intra-class edges $\mathcal{E}^{intra}$ and inter-class edges $\mathcal{E}^{inter}$ in similar ways. Given an intra-class edge $e_{i,j}\in{\mathcal{E}^{intra}}$ and the intra-class relationship feature $\mathbf{s}^{intra}$ of the expression $r$, the attention value for $e_{i,j}$ is calculated as,
\begin{align}
\begin{split}
&\mathbf{e}^{intra}_{ij} = f^{intra}_{emb}(\mathbf{e}_{ij})\\
&\mathbf{e}^{a,intra}_{ij} = \tanh(\mathbf{W}^a_{s,intra}\mathbf{s}^{intra}+\mathbf{W}^a_{g,intra}\mathbf{e}^{intra}_{ij})\\
&{A^{intra}_{ij}}'= \mathbf{w}^\intercal_{a,intra}\mathbf{e}^{a,intra}_{ij}\\
&{A^{intra}_{ij}}=\frac{\exp({A^{intra}_{ij}}')}{\sum_{k\in\mathcal{N}^{intra}_i}{\exp({A^{intra}_{ik}}')}},
\end{split}
\label{equ:intra}
\end{align}
where $f^{intra}_{emb}$ is a MLP to encode the edge feature, $\mathbf{W}^a_{g,intra}$ and $\mathbf{W}^a_{s,intra}$ map the encoded edge feature and intra-class relationship feature $\mathbf{s}^{intra}$ of expression $r$ into vectors of the same dimensionality, $\mathbf{w}_{a,intra}$ calculates the intra-class attention values for ${e_{ij}}$, and these attention values are normalised among the intra-class neighbourhood $\mathcal{N}^{intra}_i$ of $v_i$ via a softmax.

\vspace{-12pt}
\paragraph{The inter-class edge attention} The attention value for inter-class edge $e_{ij} \in{\mathcal{E}^{inter}}$ is calculated under the guidance of the inter-class relationship feature $\mathbf{s}^{inter}$ of expression $r$,
\begin{align}
\begin{split}
&\mathbf{e}^{inter}_{ij} = f^{inter}_{emb}([\mathbf{e}_{ij}, \mathbf{x}^{obj}_j])\\
&\mathbf{e}^{a,inter}_{ij} = \tanh(\mathbf{W}^a_{s,inter}\mathbf{s}^{inter}+\mathbf{W}^a_{g,inter}\mathbf{e}^{inter}_{ij})\\
&{A^{inter}_{ij}}'= \mathbf{w}^\intercal_{a,inter}\mathbf{e}^{a,inter}_{ij}\\
&{A^{inter}_{ij}}=\frac{\exp({A^{inter}_{ij}}')}{\sum_{k\in\mathcal{N}^{inter}_i}{\exp({A^{inter}_{ik}}')}},
\end{split}
\label{equ:inter}
\end{align}
where $f^{inter}_{emb}$ is a MLP. Comparing Eq.~\ref{equ:intra} and Eq.~\ref{equ:inter}, the features used to represent the intra-class relationship and inter-class relationship are different. When the subject $v_i$ and object $v_j$ are from the same category, we only use their relative spatial feature $\mathbf{e}_{ij}$ to represent the relationship between them. However, when $v_i$ and $v_j$ are from different classes ({\eg man riding horse}) we need to explicitly model the object $v_j$ and thus we design the relationship representation to be the concatenation of the edge feature $\mathbf{e}_{ij}$ and the node feature $\mathbf{x}^{obj}_j$.

%Note that there are two main differences for how to obtain the attention values for $E^r$ and $E^t$. First, as the intra-class edges represent relationships between roles of the same category (e.g., taller, largest, in the middle), the relative spatial information is sufficient to encode such relationships. Thus the attention function is applied on the edge features only. One the other hand, to identify the inter-class relationships in the form of \textit{subject+relation+object}(riding horse, throwing frisbee), we need to consider the \textit{relation} (riding) as well as the \textit{object} (horse). Due to this reason, we concatenate $\mathbf{e}_{ij}$ and $\mathbf{x}^o_j$ to represent the inter-class relationships. Second, comparing Eq.~(19) and Eq.~(23), the attention value normalization schemes are different. The intra-class relationship tends to be a relative relationship that involves multiple objects (e.g., the tallest man). To identify the edges relevant to such relationship, we use a local normalization scheme. That is, we fix an object first and then search the edge combination that can fit the relationship referred to by the expression. For example, to find ``the person between two other persons'', we need to traverse each person and see if there are people on both sides. The inter-class relationships, however, tend to describe some absolute relationship, which can be searched globally from all edges of the graph.

\begin{table*}[t!]
\renewcommand\arraystretch{1.1}
\centering
\small
\caption{Structures of MLPs. The number after linear and DP (dropout) denotes the dim of the hidden layer and the dropout ratio.}
\label{tab: mlp}
\begin{tabular}{c|c|c}
\hline
\textbf{MLPs}&\textbf{Illustration}&\textbf{Structure}\\
\hline
$f_e$& encoding one-hot representations of words in \ref{sec: lang_attn}& linear (512)+ReLU\\
$f^v_{emb}$,$f^l_{emb}$ & encoding the visual and spatial features of nodes in Eq.~\ref{equ:obj} & linear(512)+BN+ReLU+DP(0.4)+linear(512)+BN+ReLU\\
$f^{intra}_{emb}$,$f^{inter}_{emb}$ & encoding the intra and inter-class edge features in Eq.~\ref{equ:intra},~\ref{equ:inter} & linear(512)+BN+ReLU+DP(0.4)+linear(512)+BN+ReLU\\
\hline
\end{tabular}
\end{table*}

%A*** \subsubsection{Final representation for graph}
\subsubsection{The Attended Graph Representation}
\label{sec:final_representation}
With the node and edge attention determined  under the guidance of the expression $r$, the next step is to obtain the final representation for the object by aggregating the attended content. Corresponding to the decomposition of the expression, we obtain three types of features for each node: object features, intra-class relationship features, and inter-class relationship features.

The node representation for $v_i$ will be updated to $\hat{\mathbf{x}}^{obj}_i$,
\begin{align}
\vspace{-3pt}
\hat{\mathbf{x}}^{obj}_i = A^{obj}_i\mathbf{x}^{e,obj}_i,
\vspace{-3pt}
\end{align}
where $A^{obj}_i$ denotes the node attention value for $v_i$ and $\mathbf{x}^{e,obj}_i$ is the encoded node feature in Eq. \ref{equ:obj}.

The intra-class relationship representation $\hat{\mathbf{x}}^{intra}_i$ will be the weighted sum of the intra-class edge representations,
\begin{align}
\vspace{-3pt}
\hat{\mathbf{x}}^{intra}_i = \sum_{j\in\mathcal{N}^{intra}_i}A^{intra}_{ij}\mathbf{e}^{intra}_{ij},
\vspace{-3pt}
\end{align}
where $\mathcal{N}^{intra}_i$ denotes the intra-class neighbourhood of $v_i$, $A^{intra}_{ij}$ denotes the intra-class edge attention value and $\mathbf{e}^{intra}_{ij}$ is the encoded intra-class edge feature in Eq.~\ref{equ:intra}.

The inter-class relationship representation $\hat{\mathbf{x}}^{inter}_i$ is obtained as the weighted sum of the inter-class edge representations,
\begin{align}
\vspace{-3pt}
\hat{\mathbf{x}}^{inter}_i = \sum_{j\in\mathcal{N}^{inter}_i}A^{inter}_{ij}\mathbf{e}^{inter}_{ij},
\vspace{-3pt}
\end{align}
where $\mathcal{N}^{inter}_i$ denotes the inter-class neighbourhood of $v_i$, $A^{inter}_{ij}$ denotes the inter-class edge attention value and $\mathbf{e}^{inter}_{ij}$ is the encoded inter-class edge feature in Eq.~\ref{equ:inter}.

\subsection{Matching Module and Loss Function}
\label{sec: matching}
The matching score between the expression $r$ and an object $v_i$ is calculated as the weighted sum of three parts: subject, intra-class relationship, and inter-class relationship,
%To locate the object of interest, the language features in Sec.~\ref{sec: lang_attn} are compared to the object representations generated from Sec.\ref{sec:final_representation}. The matching score  between the expression $r$ and an object $o_i$ is the weighted sum of three components, where the weights are from Sec.~\ref{sec: lang_attn} and the three parts correspond to subject, intra-class relationship and inter-class relationship,
{\small
\begin{align}
\begin{split}
&p^{obj}_i = \tanh(\mathbf{W}^m_{s,subj}\mathbf{s}^{obj})^\intercal\tanh(\mathbf{W}^{m}_{g,obj}\hat{\mathbf{x}}^{obj}_i) \\
&p^{intra}_i = \tanh(\mathbf{W}^m_{s,intra}\mathbf{s}^{intra})^\intercal\tanh(\mathbf{W}^{m}_{g,intra}\hat{\mathbf{x}}^{intra}_i) \\
&p^{inter}_i = \tanh(\mathbf{W}^m_{s,inter}\mathbf{s}^{inter})^\intercal\tanh(\mathbf{W}^{m}_{g,inter}\hat{\mathbf{x}}^{inter}_i)\\
&p_i = w^{subj}p^{obj}_i+w^{intra}p^{intra}_i+w^{inter}p^{inter}_i,
\end{split}
\label{equ:match}
\end{align}}
where each expression component feature and object component feature are encoded by a MLP (linear mapping + non-linear function $\tanh(\cdot)$) respectively before a dot product.  The weights of the three parts are obtained from $r$ as introduced in Sec.~\ref{sec: lang_attn}.

The probability for $v_i$ being the referent is $prob_i=softmax(p_i)$, where the softmax is applied over all of the objects in the image. We choose CrossEntropy as the loss function. That is, if the ground truth label of $r$ is $l(r)\in[0,\cdots,N-1]$, then the loss function will be,
\begin{align}
\mathcal{L}=-\sum_{r}\log(prob_{l(r)}).
\end{align}

\section{Experiments}
\label{sec:exp}
In this section, we introduce some key implementation details,  followed by three experimental datasets. Then we present some quantitative comparisons between our method and existing works. Further, an ablation study shows the effectiveness of the key aspects of our method. Finally, visualisation for LGRANs are shown.
\subsection{Implementation details}
As mentioned in Sec.~\ref{sec: graph_construction}, we use VGG16 \cite{Simonyan14c} pre-trained on ImageNet \cite{ILSVRC15} to extract visual features for the objects in the image. In this paper, several MLPs are adopted to encode various feature representations. The details of these MLPs are illustrated in Tab.~\ref{tab: mlp}. The dimensionalities of the final representations of language representations $\{\mathbf{s}^m\}$ and object representations $\{\mathbf{x}^m_i\}$ are all 512, where $\{m\}$ denote different components. The training batch size is 30, which means in each training iteration we feed 30 images and all the referring expressions associated with these images to the network. Adam \cite{adam} is used as the training optimizer, with initial learning rate to be $0.001$, which decays by a factor of 10 every 6000 iterations. The network is implemented based on PyTorch.%\footnote{https://pytorch.org/}.

\subsection{Datasets}
We conduct experiments on three referring expression comprehension datasets: RefCOCO \cite{refcoco}, RefCOCO+ \cite{refcoco} and RefCOCOg \cite{mao2016generation}, which are all built on MSCOCO~\cite{coco}. The RefCOCO and RefCOCO+ are collected in an iterative game, where the referring expressions tend to be short phrases. The difference between these two datasets is that absolute location words are not allowed in the expressions in RefCOCO+. The expressions in RefCOCOg are longer declarative sentences. RefCOCO has 142,210 expressions for 50,000 objects in 19,994 images, RefCOCO+ has 141,565 expressions for 49,856 objects in 19,992 images, and RefCOCOg has 104,560 expressions for 54,822 objects in 26,711 images.

There are four splits for RefCOCO and RefCOCO, including ``train'', ``val'', ``testA'', ``testB''. ``testA'' and ``testB'' have different focus in evaluation. While ``testA'' has multiple persons, ``testB'' has multiple objects from other categories. For RefCOCOg, there are two data partition versions. One version is obtained by randomly splitting the objects into ``train'' and ``test''. As the data is split by objects, the same image can appear in both ``train'' and ``test''. Another partition was generated in \cite{nagaraja16refexp}. In this split, the images are split into ``train'', ``val'' and ``test''. We adopt this split for evaluation.

\subsection{Experimental results}
In this part, we show the experimental results on RefCOCO, RefCOCO+ and RefCOCOg. Accuracy is used as evaluation metric. Given an expression $r$ and a test image $I$ with a set of regions $\{o_i\}$, we use Eq.~\ref{equ:match} to select the region with highest matching score with $r$ as the prediction $o_{pred}$. Assume the referent of $r$ is $o^{\star}$, we compute the intersection-over-union (IOU) between $o_{pred}$ and $o^{\star}$ and treat the prediction correct if IOU $> 0.5$.
First, we show the comparison with state-of-the-art approaches on ground-truth MSCOCO regions. That is, for each image, the object regions $\{o_i\}$ are given. Then, we conduct ablation study to evaluate the effectiveness of two attention components and their combination, \ie node attention, edge attention and graph attention. Finally, the comparison with existing approaches on automatic detected regions are given.

\vspace{-12pt}
\paragraph{Overall Results} Tab.~\ref{tab: GT} shows the comparison between our method and state-of-the-art approaches on ground-truth regions. As can be seen, our method outperforms the other methods on almost all splits. CMN \cite{Hu_2017_CVPR} and MattNet \cite{yu2018mattnet} are relevant to our method in the sense that they abandon the monolithic language representations and use self-attention mechanism to decompose the language into different parts. However, their approaches are limited by the static and heuristic object representations, which are formed as the stack of multiple features without being informed by the expression query. We use graph attention mechanism to dynamically identify the content relevant to the language and therefore producing more discriminative object representations. ParallelAttn \cite{Zhuang_2018_CVPR} and AccumulateAttn \cite{Deng_2018_CVPR} both focus on designing attention mechanisms to highlight the informative content of the language as well as the image to achieve better grounding performance. However, they treat the objects to be isolated and fail to model the relationships between them, which turn out to be important for identifying the object of interest.

\begin{table*}[t!]
\renewcommand\arraystretch{1.2}
  \centering
  \caption{Performance (Acc\%) comparison with state-of-the-art approaches on ground-truth MSCOCO regions. ``\textbf{Speaker}+listener+reinforcer'' and ``speaker+\textbf{listener}+reinforcer'' mean using the speaker or listener module of a joint module~\cite{yu2017refexpr} to do the comprehension task respectively. All comparing methods use VGG16 features.}
   \label{tab: GT}%
   \small
    \begin{tabular}{|l|c|c|c|c|c|c|c|c|c|}
    \hline
    \multicolumn{1}{|c|}{\multirow{2}[4]{*}{Methods}} & \multicolumn{3}{c|}{RefCOCO} & \multicolumn{3}{c|}{RefCOCO+} & \multicolumn{3}{c|}{RefCOCOg} \\
    \cline{2-10}
        & val   & testA & testB & val   & testA & testB & val*  & val   & test \\
    \hline
    MMI \cite{mao2016generation}  &  -     & 71.72 & 71.09 &   -    & 58.42 & 51.23 & 62.14 &    -   &  -\\
    visdif \cite{unc_refexp} &   -    & 67.57 & 71.19 &   -    & 52.44 & 47.51 & 59.25 &   -    & - \\
    visdif+MMI \cite{unc_refexp} &    -   & 73.98 & 76.59 &   -    & 59.17 & 55.62 & 64.02 &   -    & - \\
   NegBag \cite{nagaraja16refexp} & 76.90  & 75.60  & 78.00    &   -    &   -    &   -    &   -    &   -    & 68.40 \\
    CMN \cite{Hu_2017_CVPR}  &   -    & 75.94 & 79.57 &   -    & 59.29 & 59.34 & 69.3  &  -     & - \\
    listener \cite{yu2017refexpr} & 77.48 & 76.58 & 78.94 & 60.5  & 61.39 & 58.11 & 71.12 & 69.93 & 69.03 \\
   \textbf{speaker}+listener+reinforcer \cite{yu2017refexpr} & 78.14 & 76.91 & 80.1  & 61.34 & 63.34 & 58.42 & 72.63 & 71.65 & 71.92 \\
    speaker+\textbf{listener}+reinforcer \cite{yu2017refexpr} & 78.36 & 77.97 & 79.86 & 61.33 & 63.1  & 58.19 & 72.02 & 71.32 & 71.72 \\
    VariContxt \cite{Zhang_2018_CVPR} &   -    & 78.98 & 82.39 &   -    & 62.56 & 62.90  & 73.98 &   -    & - \\
    ParallelAttn \cite{Zhuang_2018_CVPR} & 81.67 & 80.81 & 81.32 & 64.18 & 66.31 & 61.46 & 69.47 &    -   & - \\
   AccumulateAttn \cite{Deng_2018_CVPR} & 81.27 & 81.17 & 80.01 & 65.56 & \textbf{68.76} & 60.63 & 73.18 &  -     & - \\
   MattNet \cite{yu2018mattnet} & 80.94 & 79.99 & 82.3  & 63.07 & 65.04 & 61.77 & 73.08 & 73.04 & 72.79 \\
    \hline
    Ours-LGRANs & \textbf{82.0}    & \textbf{81.2}  & \textbf{84.0}    & \textbf{66.6}  & 67.6  & \textbf{65.5}  &   -    & \textbf{75.4}  & \textbf{74.7} \\
    \hline
    \end{tabular}%
  \vspace{-3pt}
\end{table*}%

\begin{table}[t!]
\renewcommand\arraystretch{1.2}
\scriptsize
  \centering\
  \vspace{-4pt}
  \caption{Ablation study of key components of LGRANs.}
  \label{tab: ablation}%
  \resizebox{1\linewidth}{!}{
    \begin{tabular}{|l|c|c|c|c|c|c|c|c|}
    \hline
    \multicolumn{1}{|c|}{\multirow{2}[4]{*}{Methods}} & \multicolumn{3}{c|}{RefCOCO} & \multicolumn{3}{c|}{RefCOCO+} & \multicolumn{2}{c|}{RefCOCOg} \\
    \cline{2-9}
          & val   & testA & testB & val   & testA & testB & val   & test \\
    \hline
    NodeRep & 77.6 & 77.7  & 77.8  & 61.5  & 62.8  & 58.0    & 67.1  & 68.4 \\
    GraphRep & 80.2 & 79.4 & 81.5 & 63.3 & 64.4 & 61.9 & 70.5 & 72.1\\
    NodeAttn & 81.4  & 80.4 & 82.8  & 65.8 & 66.2 & 64.2 & 72.4  & 73.2 \\
    EdgeAttn & 81.9  & 80.8  & 83.3  & 65.9  & 66.7  & 64.9  & 73.9  & 74.5 \\
    LGRANs & 82.0    & 81.2  & 84.0    & 66.6  & 67.6  & 65.5  & 75.4  & 74.7 \\
    \hline
    \end{tabular}}%
    \vspace{-9pt}
    
\end{table}%

\vspace{-10pt}
\paragraph{Ablation Study} Next, we conduct an ablation study to further investigate the key components of LGRANs. Specifically, we compare the following solutions:
\begin{itemize}
\item Node Representation (NodeRep): this baseline uses LSTM to encode the expression and uses
the encodings of node features to represent the objects, \ie $\mathbf{x}^{e,obj}_i$ in Eq.~\ref{equ:obj}.

\item Graph Representation (GraphRep): apart from node representation, graph representation uses two other types of edge representations: pooling of the intra-class edge features $\hat{\mathbf{x}}^{pool,intra}_i = \sum_{j\in\mathcal{N}^{intra}_i}\mathbf{e}^{intra}_{ij}$, where $\mathbf{e}^{intra}_{ij}$ is the intra-class edge feature encoding in Eq.~\ref{equ:intra}, and pooling of inter-class edge feature $\hat{\mathbf{x}}^{pool,inter}_i = \sum_{j\in\mathcal{N}^{inter}_i}\mathbf{e}^{inter}_{ij}$, where $\mathbf{e}^{inter}_{ij}$ is the inter-class edge feature encoding in Eq.~\ref{equ:inter}.

\item NodeAttn: on top of graph representation, NodeAttn applies node attention as introduced in Sec.~\ref{sec:final_representation}.

\item EdgeAttn: different from graph representation that directly aggregates the edge features, EdgeAttn applies edge attention to the edges, as introduced in Sec.~\ref{sec:final_representation}.

\item LGRANs: this is our full model, which applies both node attention and edge attention on the graph.

\end{itemize}

\begin{table}[t!]
\renewcommand\arraystretch{1.2}
  \centering
  \caption{Performance (Acc\%) comparison with state-of-the-art approaches on automatically detected regions. All comparing methods use VGG16 features.}
  \small
  \resizebox{1\linewidth}{!}{
    \begin{tabular}{|l|c|c|c|c|c|}
    \hline
    \multicolumn{1}{|c|}{\multirow{2}[4]{*}{Methods}} & \multicolumn{2}{c|}{RefCOCO} & \multicolumn{2}{c|}{RefCOCO+} & \multicolumn{1}{l|}{RefCOCOg} \\
\cline{2-6}          & testA & testB & testA & testB & val* \\
    \hline
    MMI \cite{mao2016generation}  & 64.9  & 54.51 & 54.03 & 42.81 & 45.85 \\
    NegBag \cite{nagaraja16refexp} & 58.6  & 56.4  &   -    &   -    & 39.5 \\
    CMN \cite{Hu_2017_CVPR}  & 71.03 & 65.77 & 54.32 & 47.76 & 57.47 \\
    listener \cite{yu2017refexpr} & 71.63 & 61.47 & 57.33 & 47.21 & 56.18 \\
    \textbf{spe}+lis+Rl \cite{yu2017refexpr} & 69.15 & 61.96 & 55.97 & 46.45 & 57.03 \\
    spe+\textbf{lis}+RL \cite{yu2017refexpr} & 72.65 & 62.69 & 58.68 & 48.23 & 58.32 \\
    VariContxt \cite{Zhang_2018_CVPR} & 73.33 & \textbf{67.44} & 58.40  & 53.18 & 62.30 \\
    ParallelAttn \cite{Zhuang_2018_CVPR} & 75.31 & 65.52 & 61.34 & 50.86 & 58.03 \\
    \hline
    LGRANs & \textbf{76.6}  & 66.4  & \textbf{64.0}    & \textbf{53.4} & \textbf{62.5} \\
    \hline
    \end{tabular}}%
    \vspace{-13pt}
  \label{tab: DT}%
\end{table}%

Tab.~\ref{tab: ablation} shows the ablation study results. The limitation for the baseline ``\textbf{Node Representation}'' is that it treats the objects to be isolated and ignores the relationships between objects. The ``\textbf{Graph Representation}'' considers the relationships between objects by pooling the edge features directly. It observes some improvement comparing to the baseline. ``\textbf{NodeAttn}'' takes a step further by applying the region level attention to highlight the potential object described by the expression and this further improves the performance. Orthogonal to ``NodeAttn'', ``\textbf{EdgeAttn}'' identifies the relationships relevant to the expression and this strategy results in a boost up to 3.4\%. ``\textbf{GraphAttn}'' is our full model. As can be seen, it consistently outperforms the other incomplete solutions above.

\vspace{-10pt}
\paragraph{Automatically detected regions} Finally, we evaluate the performance of LGRANs using automatically detected object regions from Faster R-CNN \cite{faster-rcnn}. Tab.~\ref{tab: DT} shows the results. Comparing to using ground-truth regions, the performance of all methods drops, which is due to the quality of the detected regions. In this setting, \textbf{LGRANs} still performs consistently better than other comparing methods. This shows the capacity of LGRANs in fully automatic referring expression comprehension. 

\begin{figure*}[t]
\begin{center}
\includegraphics[scale=.56]{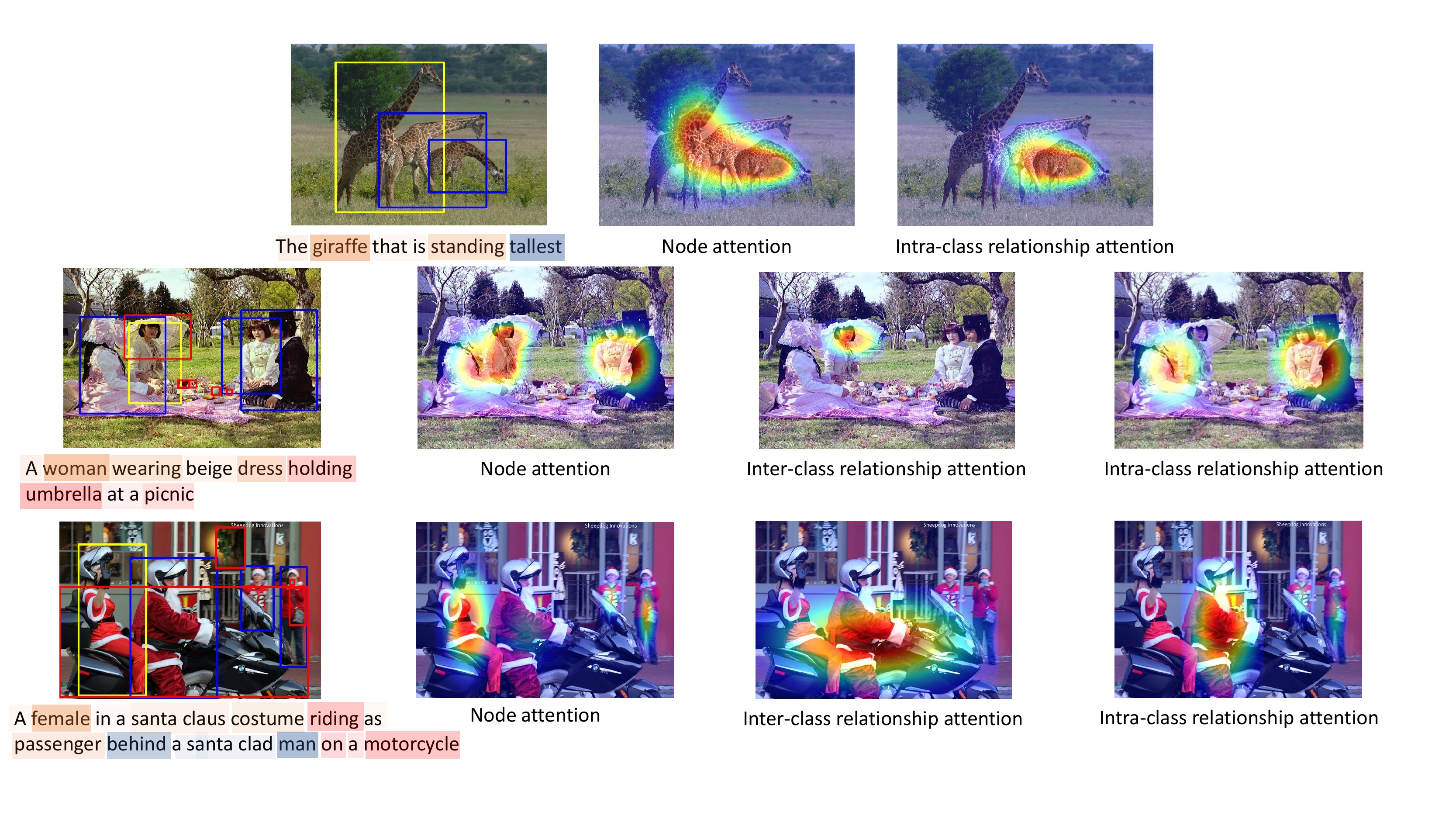}
\end{center}
\vspace{-0.3cm}
\caption{Visualisation for LGRANS. Three examples with variant difficulty levels are shown. For each example, the original image (the referent, its intra- and inter-class neighbourhood are marked by yellow, blue and red boxes respectively) with referring expression (subject, intra- and inter-class relationships are marked by yellow, blue and red), the node attention maps, the inter- and intra-class edge attention maps are given from left to right. We visualise a relationship between referent and other objects by highlighting the other objects that interact with the referent. For example, for relationship ``woman holding umbrella'', we highlight ``umbrella''.
Since the giraffe example contains giraffe regions only, no inter-class relationships exist. Within each region, the attention value is smoothed by a 2D Gaussian kernel with the centre to be the region centre. Best viewed in colour.}
\vspace{-0.3cm}
\label{fig:vis_attn}
\end{figure*}

\subsection{Visualisation}
In contrast to conventional attention schemes that apply on isolated image regions, \eg uniform grid of CNN feature maps \cite{Wang_2017_CVPR} or object proposals \cite{Anderson_2018_CVPR}, LGRANS simultaneously predict attention distributions over objects and inter-object relationships.

Fig.~\ref{fig:vis_attn} shows three examples with variant difficulty levels. In the first example, \emph{node attention} highlights all three giraffes and thus cannot distinguish the referent. To identify the tallest giraffe, it needs to compare one giraffe to the other two. As seen, the \emph{intra-class relationship attention} highlights the relationships to the other two giraffes and provides useful clues to make correct localisation. In the second example, there are four women in the image. \emph{Node attention} puts attention on people and excludes other objects, \eg bag, bottle, umbrella. Then \emph{inter-class relationship attention} identifies a relevant relationship between the referent and an umbrella. Since there are no \emph{intra-class relationships} present in the expression, the intra-class edge attention values almost evenly distribute on other persons. In the last example, the \emph{intra-class attention} and \emph{inter-class attention} identify man and motorcycle respectively, which correspond to ``behind a Santa clad man'' and ``on a motorcycle''. In these examples, the useful information present in the expression is highlighted and this explains why an object is selected as referent.

\section{Conclusion}
\label{sec:conclusion}
In this paper, we proposed a graph-based, language-guided attention networks (LGRANs) to address the referring expression comprehension task. LGRANs is composed of two key components: a \emph{node attention} component and an \emph{edge attention} component, both guided by the \emph{language attention}. The \emph{node attention} highlights the referent candidates, narrowing down the search space for localising the referent, and the \emph{edge attention} identifies the relevant relationships between the referent and its neighbourhood as informative clues. Based on the attended graph, we can dynamically enrich the object representation that better adapts to the referring expression. Another benefit of LGRANs is that it renders the comprehension decision to be visualisable and explainable.

%\clearpage
{\small
\bibliographystyle{ieee}
\bibliography{egbib}

\begin{thebibliography}{10}\itemsep=-1pt

\bibitem{Anderson_2018_CVPR}
P.~Anderson, X.~He, C.~Buehler, D.~Teney, M.~Johnson, S.~Gould, and L.~Zhang.
\newblock Bottom-up and top-down attention for image captioning and visual
  question answering.
\newblock In {\em {Proc. IEEE Conf. Comp. Vis. Patt. Recogn.}}, 2018.

\bibitem{7780381}
J.~Andreas, M.~Rohrbach, T.~Darrell, and D.~Klein.
\newblock Neural module networks.
\newblock In {\em {Proc. IEEE Conf. Comp. Vis. Patt. Recogn.}}, 2016.

\bibitem{Deng_2018_CVPR}
C.~Deng, Q.~Wu, Q.~Wu, F.~Hu, F.~Lyu, and M.~Tan.
\newblock Visual grounding via accumulated attention.
\newblock In {\em {Proc. IEEE Conf. Comp. Vis. Patt. Recogn.}}, 2018.

\bibitem{CNG}
D.~Duvenaud, D.~Maclaurin, J.~Aguilera-Iparraguirre, R.~G\'{o}mez-Bombarelli,
  T.~Hirzel, A.~Aspuru-Guzik, and R.~P. Adams.
\newblock Convolutional networks on graphs for learning molecular fingerprints.
\newblock In {\em {Proc. Advances in Neural Inf. Process. Syst.}}, 2015.

\bibitem{Hu2017LearningTR}
R.~Hu, J.~Andreas, M.~Rohrbach, T.~Darrell, and K.~Saenko.
\newblock Learning to reason: End-to-end module networks for visual question
  answering.
\newblock {\em {Proc. IEEE Int. Conf. Comp. Vis.}}, 2017.

\bibitem{Hu_2017_CVPR}
R.~Hu, M.~Rohrbach, J.~Andreas, T.~Darrell, and K.~Saenko.
\newblock Modeling relationships in referential expressions with compositional
  modular networks.
\newblock In {\em {Proc. IEEE Conf. Comp. Vis. Patt. Recogn.}}, 2017.

\bibitem{Hu2016NaturalLO}
R.~Hu, H.~Xu, M.~Rohrbach, J.~Feng, K.~Saenko, and T.~Darrell.
\newblock Natural language object retrieval.
\newblock In {\em CVPR}, 2016.

\bibitem{Jain_2016_CVPR}
A.~Jain, A.~R. Zamir, S.~Savarese, and A.~Saxena.
\newblock Structural-rnn: Deep learning on spatio-temporal graphs.
\newblock In {\em {Proc. IEEE Conf. Comp. Vis. Patt. Recogn.}}, 2016.

\bibitem{refcoco}
S.~Kazemzadeh, V.~Ordonez, M.~Matten, and T.~L. Berg.
\newblock Referit game: Referring to objects in photographs of natural scenes.
\newblock In {\em {Proc. Conf. Empirical Methods in Natural Language
  Processing}}, 2014.

\bibitem{adam}
D.~P. Kingma and J.~Ba.
\newblock Adam: A method for stochastic optimization.
\newblock In {\em {Proc. Int. Conf. Learn. Representations}}, 2014.

\bibitem{SituationsICCV17}
R.~Li, M.~Tapaswi, R.~Liao, J.~Jia, R.~Urtasun, and S.~Fidler.
\newblock Situation recognition with graph neural networks.
\newblock In {\em {Proc. IEEE Int. Conf. Comp. Vis.}}, 2017.

\bibitem{grn}
Y.~Li, D.~Tarlow, M.~Brockschmidt, and R.~S. Zemel.
\newblock Gated graph sequence neural networks.
\newblock In {\em {Proc. Int. Conf. Learn. Representations}}, 2016.

\bibitem{coco}
T.-Y. Lin, M.~Maire, S.~Belongie, J.~Hays, P.~Perona, D.~Ramanan,
  P.~Doll{\'a}r, and C.~L. Zitnick.
\newblock Microsoft coco: Common objects in context.
\newblock In {\em {Proc. Eur. Conf. Comp. Vis.}}, 2014.

\bibitem{Luo2017}
R.~Luo and G.~Shakhnarovich.
\newblock Comprehension-guided referring expressions.
\newblock {\em {Proc. IEEE Conf. Comp. Vis. Patt. Recogn.}}, 2017.

\bibitem{mao2016generation}
J.~Mao, J.~Huang, A.~Toshev, O.~Camburu, A.~Yuille, and K.~Murphy.
\newblock Generation and comprehension of unambiguous object descriptions.
\newblock In {\em {Proc. IEEE Conf. Comp. Vis. Patt. Recogn.}}, 2016.

\bibitem{nagaraja16refexp}
V.~K. Nagaraja, V.~I. Morariu, and L.~S. Davis.
\newblock Modeling context between objects for referring expression
  understanding.
\newblock In {\em {Proc. Eur. Conf. Comp. Vis.}}, 2016.

\bibitem{faster-rcnn}
S.~Ren, K.~He, R.~Girshick, and J.~Sun.
\newblock Faster r-cnn: Towards real-time object detection with region proposal
  networks.
\newblock In {\em {Proc. Advances in Neural Inf. Process. Syst.}}, 2015.

\bibitem{reconstruction}
A.~Rohrbach, M.~Rohrbach, R.~Hu, T.~Darrell, and B.~Schiele.
\newblock Grounding of textual phrases in images by reconstruction.
\newblock In {\em {Proc. Eur. Conf. Comp. Vis.}}, 2016.

\bibitem{ILSVRC15}
O.~Russakovsky, J.~Deng, H.~Su, J.~Krause, S.~Satheesh, S.~Ma, Z.~Huang,
  A.~Karpathy, A.~Khosla, M.~Bernstein, A.~C. Berg, and L.~Fei-Fei.
\newblock {ImageNet Large Scale Visual Recognition Challenge}.
\newblock {\em {Int. J. Comput. Vision}}, 2015.

\bibitem{bi-lstm}
M.~Schuster and K.~Paliwal.
\newblock Bidirectional recurrent neural networks.
\newblock {\em IEEE Transactions on Signal Processing}, 1997.

\bibitem{Simonyan14c}
K.~Simonyan and A.~Zisserman.
\newblock Very deep convolutional networks for large-scale image recognition.
\newblock {\em CoRR}, abs/1409.1556, 2014.

\bibitem{Teney_2017_CVPR}
D.~Teney, L.~Liu, and A.~van~den Hengel.
\newblock Graph-structured representations for visual question answering.
\newblock In {\em {Proc. IEEE Conf. Comp. Vis. Patt. Recogn.}}, 2017.

\bibitem{Uijlings13}
J.~Uijlings, K.~van~de Sande, T.~Gevers, and A.~Smeulders.
\newblock Selective search for object recognition.
\newblock {\em {Int. J. Comput. Vision}}, 2013.

\bibitem{velickovic2018graph}
P.~Veličković, G.~Cucurull, A.~Casanova, A.~Romero, P.~Liò, and Y.~Bengio.
\newblock Graph attention networks.
\newblock In {\em {Proc. Int. Conf. Learn. Representations}}, 2018.

\bibitem{Wang_2016_CVPR}
L.~Wang, Y.~Li, and S.~Lazebnik.
\newblock Learning deep structure-preserving image-text embeddings.
\newblock In {\em {Proc. IEEE Conf. Comp. Vis. Patt. Recogn.}}, 2016.

\bibitem{Wang_2017_CVPR}
P.~Wang, L.~Liu, C.~Shen, Z.~Huang, A.~van~den Hengel, and H.~Tao~Shen.
\newblock Multi-attention network for one shot learning.
\newblock In {\em {Proc. IEEE Conf. Comp. Vis. Patt. Recogn.}}, 2017.

\bibitem{yu2018mattnet}
L.~Yu, Z.~Lin, X.~Shen, J.~Yang, X.~Lu, M.~Bansal, and T.~L. Berg.
\newblock Mattnet: Modular attention network for referring expression
  comprehension.
\newblock In {\em {Proc. IEEE Conf. Comp. Vis. Patt. Recogn.}}, 2018.

\bibitem{unc_refexp}
L.~Yu, P.~Poirson, S.~Yang, A.~C. Berg, and T.~L. Berg.
\newblock Modeling context in referring expressions.
\newblock In {\em {Proc. Eur. Conf. Comp. Vis.}}, 2016.

\bibitem{yu2017refexpr}
L.~Yu, H.~Tan, M.~Bansal, and T.~L. Berg.
\newblock A joint speaker-listener-reinforcer model for referring expressions.
\newblock In {\em {Proc. IEEE Conf. Comp. Vis. Patt. Recogn.}}, 2017.

\bibitem{Zhang_2018_CVPR}
H.~Zhang, Y.~Niu, and S.-F. Chang.
\newblock Grounding referring expressions in images by variational context.
\newblock In {\em {Proc. IEEE Conf. Comp. Vis. Patt. Recogn.}}, 2018.

\bibitem{Zhuang_2017_ICCV}
B.~Zhuang, L.~Liu, C.~Shen, and I.~Reid.
\newblock Towards context-aware interaction recognition for visual relationship
  detection.
\newblock In {\em {Proc. IEEE Int. Conf. Comp. Vis.}}, 2017.

\bibitem{Zhuang_2018_CVPR}
B.~Zhuang, Q.~Wu, C.~Shen, I.~Reid, and A.~van~den Hengel.
\newblock Parallel attention: A unified framework for visual object discovery
  through dialogs and queries.
\newblock In {\em {Proc. IEEE Conf. Comp. Vis. Patt. Recogn.}}, 2018.

\bibitem{10.1007/978-3-319-10602-1_26}
C.~L. Zitnick and P.~Doll{\'a}r.
\newblock Edge boxes: Locating object proposals from edges.
\newblock In {\em {Proc. Eur. Conf. Comp. Vis.}}, 2014.

\end{thebibliography}
}

\end{document}